\documentclass[runningheads]{llncs}
\usepackage[T1]{fontenc}
\usepackage[dvipsnames]{xcolor}
\usepackage{graphicx}
\usepackage{multirow}
\usepackage{subfig}
\usepackage{amsmath}
\usepackage{amsfonts}
\usepackage{multirow, tabularx}
\usepackage[colorlinks=true,linkcolor=red, citecolor=blue]{hyperref}%
\begin{document}

\title{Local and Global Contextual Features Fusion for Pedestrian Intention Prediction}

\author{Mohsen Azarmi\inst{}%\orcidID{0000-0003-0737-9204} 
\and
Mahdi Rezaei\inst{}%\orcidID{0000-0003-3892-421X} 
\and Tanveer Hussain \inst{} \and Chenghao Qian\inst{}}
\authorrunning{M. Azarmi et al.}
\institute{\textit{Institute for Transport Studies, University of Leeds, LS2 9JT, United Kingdom} \email{m.rezaei@leeds.ac.uk}}

\authorrunning{M. Azarmi et al.}% Part of LEFT running header
\titlerunning{Local and Global Features Fusion for
Pedestrian Intention Prediction}%

\maketitle              % typeset the header of the contribution
\begin{abstract}
Autonomous vehicles (AVs) are becoming an indispensable part of future transportation. However, safety challenges and lack of reliability limit their real-world deployment. Towards boosting the appearance of AVs on the roads, the interaction of AVs with pedestrians including ``prediction of the pedestrian crossing intention'' deserves extensive research. This is a highly challenging task as involves multiple non-linear parameters. In this direction, we extract and analyse spatio-temporal visual features of both pedestrian and traffic contexts. The pedestrian features include body pose and local context features that represent the pedestrian's behaviour. Additionally, to understand the global context, we utilise location, motion, and environmental information using scene parsing technology that represents the pedestrian's surroundings, and may affect the pedestrian's intention. Finally, these multi-modality features are intelligently fused for effective intention prediction learning. The experimental results of the proposed model on the JAAD dataset show a superior result on the combined AUC and F1-score compared to the state-of-the-art.

\end{abstract}

\section{Introduction}

Pedestrian crossing intention prediction or \textit{Pedestrian Intention Prediction} (PIP) is deemed to be important in the context of autonomous driving systems (ADS). During the past decade, a variety of approaches have investigated similar challenging tasks and recently more studies have been conducted about pedestrian crossing behaviours \cite{ridel2018literature,tian2022explaining} in the computer vision community.

This includes interpreting the upcoming actions of pedestrians which involve a high degree of freedom and complexity in their movements \cite{serpush2021complex}. 
Pedestrians can select any trajectories and might show very agile behaviours or change their motion direction abruptly. 
Pedestrians may not follow designated crossing areas or zebra-crossing  \cite{rezaei2021traffic}, and also be distracted by talking or texting over a phone or with other accompanying pedestrians. 
Their intention for crossing could also be affected by many other factors such as traffic density, demographics, walking in a group or alone, road width, road structure, and many more \cite{Schneemann2016Context}.
All these factors form contextual data for PIP. However, most of the studies such as \cite{rasouli2020pedestrian,Yang2021CrossingON,sharma2022pedestrian} have tried to investigate the relationship between only one or two of these factors and pedestrian crossing behaviour. 

AVs consider a conservative approach in interaction with pedestrians as the most vulnerable road users (VRUs), by driving at a slow pace, avoiding complex interactions, and stopping often to avoid any road catastrophe.
There is also a preference to drive in less complicated environments in terms of understanding VRUs, which limits the AVs with a low level of autonomy to participate in numerous everyday traffic scenarios \cite{wang2021towards}.
However, high levels of autonomy (i.e., levels 4 and 5) demand a higher level of interaction with VRUs. A robust PIP model can provide the information needed to realise what exactly a pedestrian is about to do in a particular traffic scenario.
Even for level 3 (conditional automation) vehicles, a PIP model buys more time for the vehicle and/or driver to take decisions and leads to a safer manoeuvre. 

Action prediction models \cite{kong2022human}, analyse current and past video sequences of a human (temporal analysis), to estimate the human action in the forthcoming frames, or for pedestrian motion estimation.
Algorithms that utilise contextual information from the scene, road infrastructure, and behavioural characteristics (e.g. pose) perform better in understanding the 
pedestrian's intention \cite{rasouli2020pedestrian}.
Temporal analysis methods rely on the past moving trajectories of pedestrians to anticipate future action of whether a pedestrian crosses or not \cite{alahi2016social}.

This study aims to consider different sources of features that could be extracted from sequential video data and pedestrian pose to develop a pedestrian crossing intention prediction model %.
using vision-based convolutional neural networks. 
The model uses spatio-temporal features, pose, and contextual information extracted from the front-view camera in JAAD dataset \cite{rasouli2017ICCVW}.

\section{Related Works}\label{related}

Pedestrian intention prediction (PIP) requires accurate detection \cite{zaidi2022survey}, tracking \cite{chen2022visual}, localisation \cite{cao2021handcrafted}, and moving trajectory estimation \cite{korbmacher2022review}. With the development of high-resolution sensors, PIP research has become more feasible and attractive. 

The study by Rasouli and Toostsos \cite{rasouli2019autonomous} conduct the interaction between AVs and pedestrians in traffic scenes. They emphasise the importance of AVs' communication with other road users, and they point out the major practical and theoretical challenges of this subject.
A set of other studies propose deep learning-based approaches for  action recognition \cite{serpush2021complex} using multi-task learning to predict the crossing intention of pedestrians \cite{pop2019multi,bouhsain2020pedestrian}. 
Other studies such as \cite{minguez2018pedestrian}, considered both moving trajectories and pedestrian' body pose with promising results in classifying pedestrian crossing behaviour.  
Although they can find an apparent relationship between body gestures and the tendency to cross, insufficient consideration of spatial features prevents them from precise intention prediction in the specialised datasets and benchmarks \cite{kotseruba2021benchmark}.

Another research \cite{rasouli2020role} uses contextual information by extending the previous work by \cite{Schneemann2016Context}, which suggests that the crossing intention ought to be considered as a context-aware problem. They suggest considering not only the pedestrian features but also the road semantic information. 
 
Several deep learning methods, such as 2D \cite{razali2021pedestrian} and 3D convolutional neural networks \cite{Yang2021CrossingON,jiang2022two}, Long-Sort Term Memory (LSTM) \cite{saleh2019real,quan2021holistic}, attention mechanism \cite{liu2020spatiotemporal}, Transformer \cite{Lorenzo2021CAPformerPC}, and graph neural networks \cite{liu2020spatiotemporal,chen2021visual}  have been utilised to assess the spatio-temporal features and to correlate pedestrian features with semantic information. 

The outcome of more successful research works shows that using multiple 
sources of information, such as different sensors \cite{zhao2019trajectory}, moving trajectories \cite{saleh2018intent}, body gesture \cite{gesnouin2020predicting,piccoli2020fussi,singh2021multi}, semantic segmentation \cite{Yang2021PredictingPC}, dynamic and motion information \cite{neogi2020context,neumann2021pedestrian} would lead to more accurate results. 
Hence, some of the recent research directions have been focused on finding optimal feature fusion strategies \cite{rasouli2020pedestrian} but are still in their infancy.

As a new contribution, this study aims to enhance the fusion approaches by incorporating environmental features and camera motion as global features and pedestrian attributes as local features to classify and predict the crossing intention. This is evaluated against the challenging dataset of \textit{Joint Attention in Autonomous Vehicle (JAAD)} \cite{rasouli2017ICCVW}.

\section{The Proposed Methodology}\label{method}

The proposed method analyses sequential video information, the global context of the scene, and the pose of the pedestrian as the input features, and as a result, predicts the final pedestrian's crossing intention or action.

We define the final action of a pedestrian as a binary classification problem consisting of two classes of ``crossing'' or ``not crossing''.

\begin{figure}[t!]
	\centering
	\includegraphics[width = 0.95\linewidth]{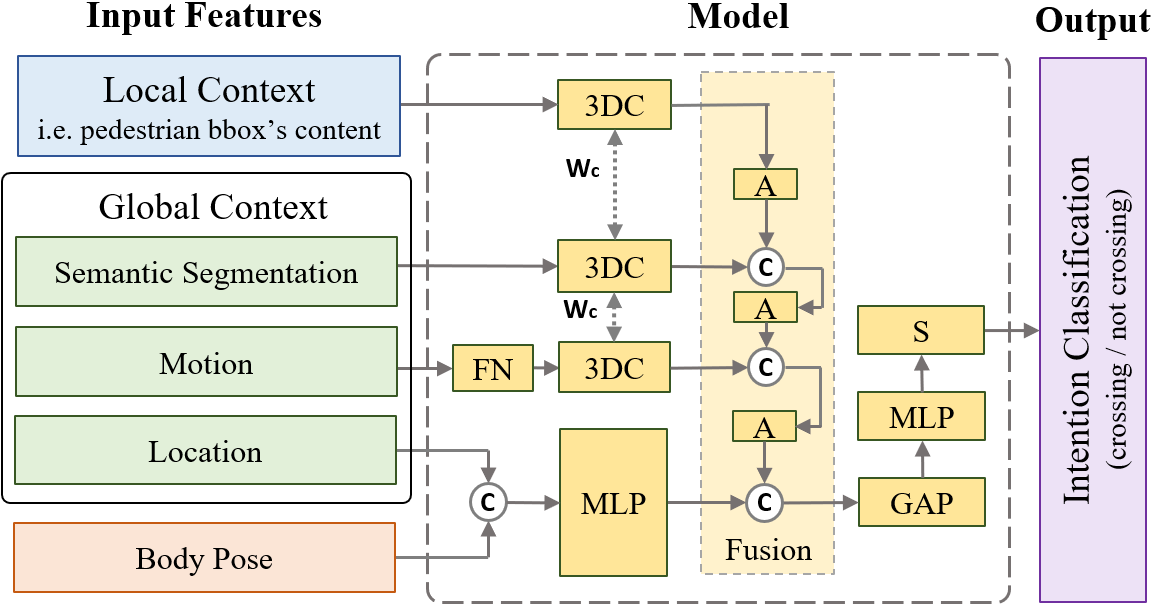}
	\caption{The overview of the proposed model. Abbreviations: \textbf{3DC} is the 3D convolutional layer;  \textbf{$W_c$} is the shared weights of the network; \textbf{MLP} is multi-layer perceptron; \textbf{FN} is Flownet2 model; \textbf{A} is self-attention module; \textbf{GAP} is global average pooling; \textbf{S} is Softmax function, and \textbf{C} refers to the feature concatenation over source dimension.}
	\label{model}
\end{figure}

Figure~\ref{model} demonstrates the model architecture and its components, which intakes the extracted features from the video footage and generates the intention classification result as the output. The proposed architecture and the types of features are explained in subsequent sections.

\subsection{Global Context}
\subsubsection{Semantic Segmentation:}
In order to enrich our model with global information surrounding the pedestrian, we parse the scene with a semantic segmentation algorithm which also 
reduces the noise of perceiving spatial information in addition to segmenting various classes of objects in the scene. 

The exploited segmentator ($S$) interprets the input image $I_{RGB(h \times w)}$, where $h, w$ are the height and width of the image, respectively, to produce output image  $I_s(h \times w \times n)$ = $S (I_{RGB(h \times w)})$, which contains $n$ binary masks and each one refers to the existence of an individual class in the scene.
The scene is categorised into eight classes of objects including pedestrians, roads, vehicles, constructions, vegetation, sky, buildings, and unrecognised objects, using the model proposed by \cite{wang2022internimage}.

\subsubsection{Motion:}
We consider dense optical flow as a scene motion descriptor ($F$) to obtain velocity measures of the pedestrians in the dataset. 
However, we avoid using the classical optical flow method which compares the intensities within a given window. Instead, we utilise a deep learning-based technology, Flownet2 \cite{ilg2017flownet}, which leverages the accuracy and accelerates the run-time performance.
The network produces the output  as $I_o(h \times w) = F(I_{RGB(h\times w)})$.

\subsubsection{Location:}
Like other state-of-the-art models \cite{liu2020spatiotemporal,chen2021visual,jiang2022two}, 
we extract the location of the pedestrians directly from the JAAD ground truth; 
although this can be done by a detection algorithm such as \cite{zaidi2022survey}.
JAAD's annotation provides the corresponding coordinates of the top-left and bottom-right corners of the bounding box.
To avoid varying scales of input data, we adopt \textit{min-max} scaling approach which normalises the bounding box coordinate values between 0 to 1 as follows:
\begin{equation}\label{scale}
    \chi = \frac{\chi - \chi_{min}}{\chi_{max} - \chi_{min}},
\end{equation}

\noindent where $x$ and $y$ value ranges from 0 to 1920 and 0 to 1080, respectively. Due to the negligible variation of width ($w_{b}$) for bounding boxes in the dataset, we follow \cite{Lorenzo2021CAPformerPC} study and remove $w_{b}$, so, the arrangement of coordinates is considered as $(x, y, h_{b})$, where the first and second elements are the centre coordinates of the bounding box, respectively, and $h_{b}$ is denoted as the height of the bounding box.
Thereafter, the coordinates input vector containing the location of the pedestrian is defined as $v_b \in \mathbb{R}^{B \times N \times 3}$, where $B$ is the batch size and $N$ is the length of the sequence of input data.

\subsection{Local Context}
Robust features of pedestrians including bounding box and pose information play a key role in designing a generalised model and should not be neglected. This gives the model the capability of producing accurate and precise predictions \cite{Mordan2020Detecting3P}.

We define ``Local context'' as a region that constitutes the content of  the pedestrian image as the shape $I_l \in \mathbb{R}^{B \times N \times h_{b} \times  w_{b} \times 3}$, where 3 is the number of channels for the red, green, and blue intensities, ranging $[0, 255]$. We also normalise this vector between 0 and 1 values using Eq. \ref{scale}

\subsection{Body Pose}
The 2D coordinates of pose key points are obtained as the output of a pose estimator algorithm as used in \cite{piccoli2020fussi}. 
The pose information includes 36 values corresponding to 18 pairs of 2D coordinates, representing the body joints of the pedestrian.
We have also normalised them using \textit{min–max} scaling (Eq. \ref{scale}) which results in values between 0 and 1. 
The vector containing the body pose is defined as $v_p \in \mathbb{R}^{B \times N \times 36}$.

\subsection{Model Architecture}
As shown in Figure \ref{model},  the model architecture consists of different components. 
We adopt three layers of a multi-layer perceptron network ($\textbf{N}_{MLP}$) to obtain the embedding of bounding box vectors ($v_b$) and body pose vector ($v_b$). We also adopt a 3D convolution network ($\textbf{N}_{3DC}$) \cite{tran2015learning} to extract spatio-temporal features of local context region, global semantic, and motion information as follows:

\begin{equation}
    \begin{aligned}
        f_b = \textbf{N}_{MLP}(v_b \oplus v_p, {\textbf{W}_{b}}),\\
        f_{c_l} = \textbf{N}_{3DC}(I_l, \textbf{W}_c),\\
        f_{c_g} = \textbf{N}_{3DC}(I_g, \textbf{W}_c),\\
        f_{c_o} = \textbf{N}_{3DC}(I_o, \textbf{W}_c),\\
    \end{aligned}
\end{equation}
where $\textbf{W}_b$ is the weight of the 3-layer MLP, and $\textbf{W}_c$ is the shared weight of the local, global, and motion features.
$f_{b}$ and $f_{c_i | i \in \{l, g, o\}}$ have the same feature vectors size of 128, where $f_b$ is dedicated to storing the pedestrian location and pose, and $f_{cl}$, $f_{cg}$, $f_{co}$ are dedicated to storing the local context information, semantic segmentation, and optical flow (motion features) of the sequential input frames, respectively.

\begin{figure}[t!]
    \centering
    \subfloat[A sample scenario where a pedestrian has an intention to cross the road]{
        \includegraphics[width = 0.95\linewidth]{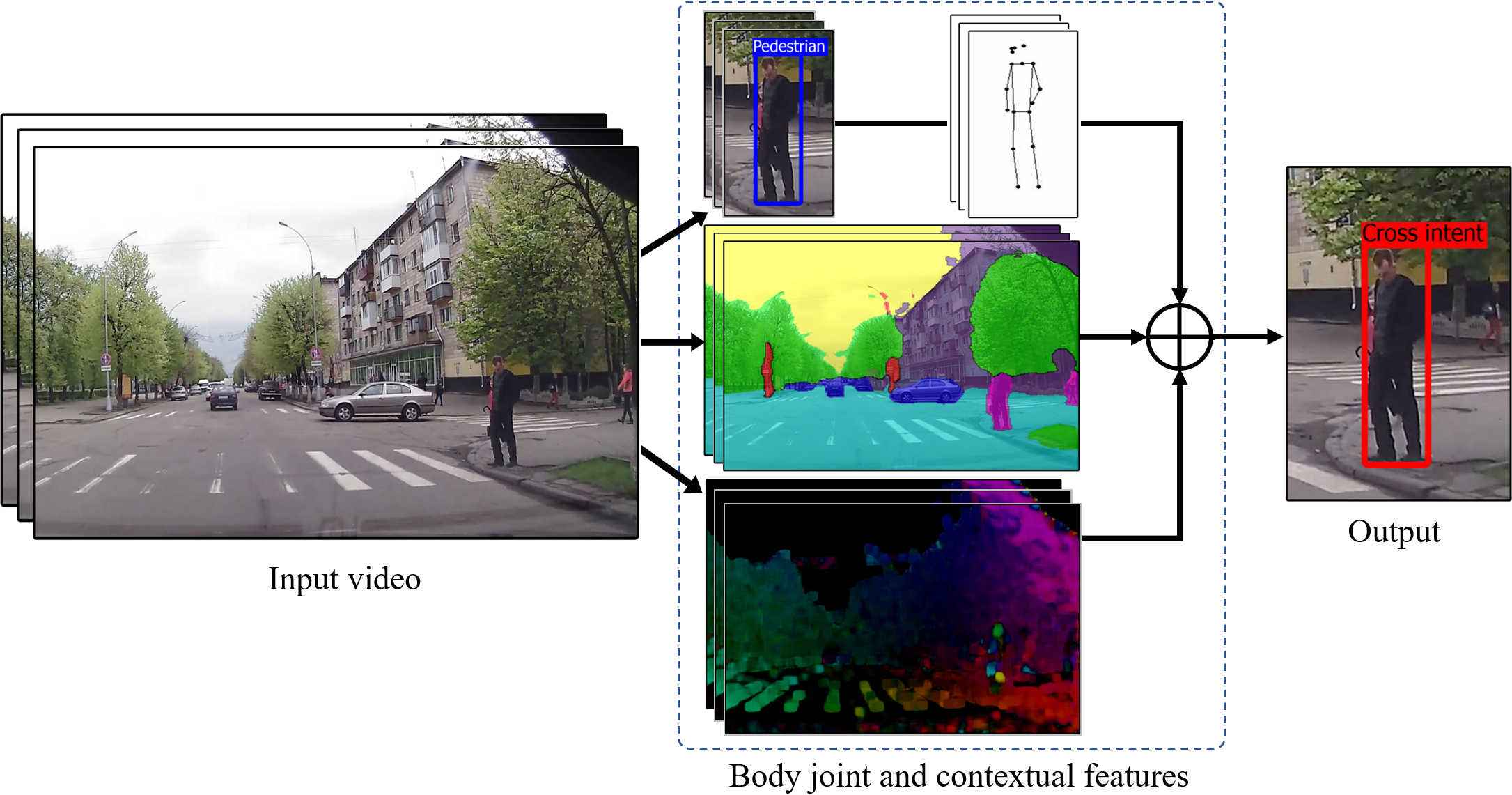}\label{out1}}\vspace{1mm}
    \subfloat[A sample scenario where a pedestrian has no intent to cross the road]{
        \includegraphics[width = 0.95\linewidth]{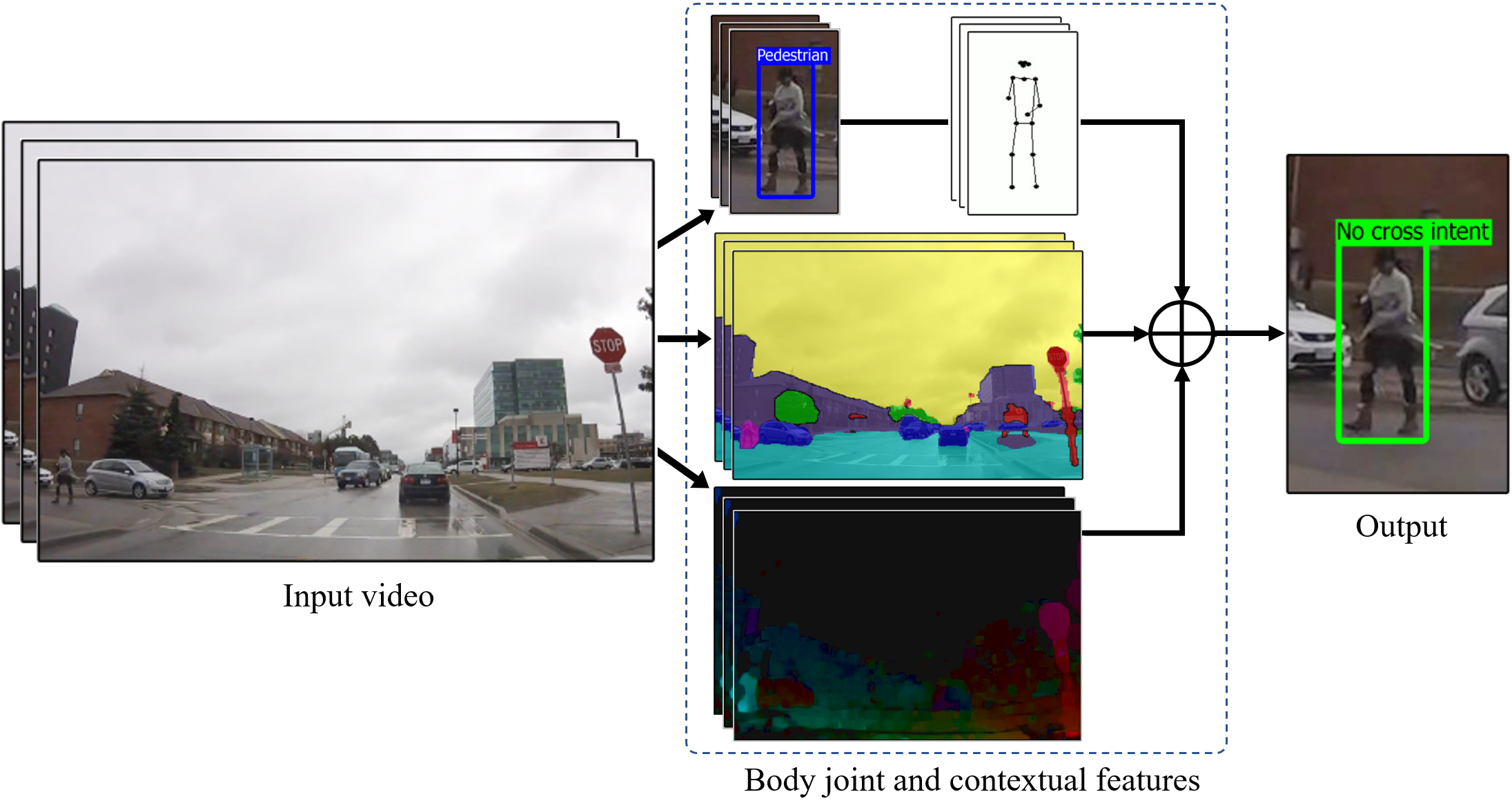}\label{out2}}\vspace{-1mm}
    
    \caption{The results of the proposed model in predicting pedestrian intention. Pedestrian local context and body pose,  location, the scene motion, and environmental information as global context information are fused to reason about the intention of the pedestrian to cross or not cross the road.}\label{results}
\end{figure}

Figure \ref{results} visualises the model flow from input to extracted features followed by crossing intention classification. Our model utilises multiple features; however, the importance factor of each feature might be different for pedestrian crossing classification purposes. 
Hence, in order to select efficient multi-modal features from a local to a global perspective, the self-attention module \cite{zhao2020exploring} ($\textbf{A}$) is applied in each fusion step.

The location and pose feature vector of the pedestrian $f_{b}^t$ at time $t$ is built based on the extracted temporal attention feature $f_A$ over the past $N$ frames. This helps to understand the importance factor of each feature. Thus, we take the learnable self-attention module, as follows:

\begin{equation}
    f_a = \textbf{A}([f_{b}^1 ; ...; f_{b}^t; ...; f_{b}^N]),
\end{equation}
where $[;]$ represents the stack concatenation over temporal dimensions.
Afterwards, the fusion of local context features, global context, motion, and the combined location and pose feature vector is specified as:

\begin{equation}
    f_{H} = \textbf{A}([f_a; \textbf{A}([f_{c_o}; \textbf{A}([f_{c_g}; f_{c_l}])])]),
\end{equation}
Similarly, here $[;]$ denotes the concatenation over the feature modal dimension, and $f_{H} \in \mathbb{R}^{128 \times K}$, where $K$ denotes the number of feature models and is set as 3 in our fusion strategy. 
Then, this vector is fed to a global average pooling (GAP) layer to enforce correspondences between feature maps and categories, to be judged by a binary classifier. The classifier consists of two fully-connected layers and a \textit{Softmax} layer. The \textit{Softmax} layer determines the probability of crossing or not crossing intention within the range of 0 to 1.

\section{Experiments}\label{experiments}
In this section, we discuss the JAAD dataset, the model hyperparameters, and pre-processing settings. We also review the performance of the model compared with similar methods. All experiments are conducted on a PC workstation with an Intel~\copyright\, Xeon W-2225 4.10 GHz processor and dual NVIDIA RTX 5000 GPU with CUDA version~12.0 to benefit the unified shared memory and accelerate the training and inference phases using parallel computing.

\subsection{Dataset}
The JAAD dataset \cite{rasouli2017ICCVW} offers 346 high-resolution videos of autonomous driving in everyday urban areas with explicit crossing behaviour annotation. The length of videos ranges from 1 second to 31 seconds, and they may contain multiple pedestrians in the scene. 
The ground truth labels on 2D locations for each frame are provided.

The dataset has two subsets, behavioural data (JAAD$_{beh}$), which contains pedestrians who are crossing (495 samples) or those who are about to cross (191 samples), and all data (JAAD$_{all}$), which has additional pedestrians (2100 samples) with non-crossing actions. To ensure a fair comparison and benchmark, we consider the same training/test splits as per \cite{kotseruba2021benchmark}.

\subsection{Implementation Details}
The training data includes 250 videos with a total number of 460 pedestrians on the scenes, and the test data includes 96 videos with a total number of 253 pedestrians in all scenes.

The distribution of labels is roughly balanced throughout the dataset, with 44.30\% frames labelled as crossing and 55.70\% as non-crossing.

As part of the hyperparameters setting, a learning rate of $5 \times 10^{-7}$, 40 epochs, AdamW~\cite{loshchilov2017decoupled} optimiser, and the batch size ($B$)~=~2 were applied. The weight decay of AdamW was deactivated, and a constant decay rate of $10^{-4}$ in fully-connected layers and a dropout of 0.5 in the attention modules were applied for regularisation and to avoid over-fitting.

As per the previous research and suggestions by \cite{liu2020spatiotemporal}, a fixed-length sequence of $N$=16 frames is used for spatio-temporal analysis using JAAD dataset. Given the pedestrian intention (action) commences at TTE=0 (time to event), the training sequences will be extracted from the past 1-2 seconds, i.e. \textit{fr}$_{N-1} \in [TTE - 60, TTE - 30]$. For those videos which contain multiple pedestrians on the scene, the training phase will be repeated multiple times to ensure most of the training samples are utilised.

In terms of data augmentation, random horizontal flip, roll rotation, and colour jittering with a probability of 50\% are applied; the same change is performed in all images, bounding boxes, and poses of a sequence. 

\subsection{Quantitative and Qualitative Evaluation}
In our experiments, the proposed model was evaluated against the JAAD ground truth, and compared with other state-of-the-art models summarised in \cite{kotseruba2021benchmark}. 

Table~\ref{exp} shows the quantitative results of the comparison in terms of area under curve (AUC) and F1 score metrics. 
AUC represents the ability of our classifier to distinguish between both classes of 'crossing' or 'not-crossing'. For example, a value of 0.5 means that classifier behaviour is equivalent to randomly choosing the class. 
F1 score is the harmonic mean of precision and recall which are well-known criteria for assessing machine learning-based classifiers.  

\begin{table}[t!]
\caption{Performance comparison between the proposed model and other models on the benchmark JAAD dataset. Various variations of our progressive study show the effect of adding local context, global contexts, and pose features. 
B refers to the bounding box, P is the body pose, L is the local context,  G refers to the global contexts, and M refers to motion information. The \textcolor{OrangeRed}{\textbf{red}} and \textcolor{ForestGreen}{\textbf{green}} numbers represent the best and second-best results for each column.} \vspace{2mm}
\centering
\begin{tabular}{l|c|cc|cc}
\hline
\multicolumn{1}{c|}{\multirow{2}{*}{Model}} & \multirow{2}{*}{Variant} & \multicolumn{2}{c|}{JAAD$_{beh}$} & \multicolumn{2}{c}{JAAD$_{all}$} \\
\multicolumn{1}{c|}{} &  & AUC & F1 & AUC & F1 \\ \hline
\multirow{2}{*}{Static} & VGG16 & 0.52 & 0.71 & 0.75 & 0.55 \\
 & ResNet50 & 0.45 & 0.54 & 0.72 & 0.52 \\
\multirow{2}{*}{ConvLSTM \cite{shi2015convolutional}} & VGG16 & 0.49 & 0.64 & 0.57 & 0.32 \\
 & ResNet50 & 0.55 & 0.69 & 0.58 & 0.33 \\
\multirow{2}{*}{SingleRNN \cite{kotseruba2020they}} & GRU & 0.54 & 0.67 & 0.59 & 0.34 \\
 & LSTM & 0.48 & 0.61 & 0.75 & 0.54 \\
\multirow{2}{*}{I3D \cite{bhattacharyya2018long}} & RGB & 0.56 & 0.73 & 0.74 & 0.63 \\
 & Optical Flow & 0.51 & 0.75 & 0.80 & 0.63 \\
C3D \cite{tran2015learning} & RGB & 0.51 & 0.75 & 0.81 & 0.65 \\
ATGC \cite{rasouli2017ICCVW} & AlexNet & 0.41 & 0.62 & 0.62 & \color{OrangeRed}{\textbf{0.76}} \\
MultiRNN \cite{bhattacharyya2018long} & GRU & 0.50 & 0.74 & 0.79 & 0.58 \\
StackedRNN \cite{yue2015beyond} & GRU & \color{OrangeRed}{\textbf{0.60}} & 0.66 & 0.79 & 0.58 \\
HRNN \cite{du2015hierarchical} & GRU & 0.50 & 0.63 & 0.79 & 0.59 \\
SFRNN \cite{rasouli2020pedestrian} & GRU & 0.45 & 0.63 & 0.84 & 0.65 \\
TwoStream \cite{simonyan2014two} & VGG16 & 0.52 & 0.66 & 0.69 & 0.43 \\
PCPA \cite{kotseruba2021benchmark} & C3D & 0.50 & 0.71 & \color{OrangeRed}{\textbf{0.87}} & 0.68 \\
CAPformer \cite{Lorenzo2021CAPformerPC} & Timesformer & 0.55 & \color{OrangeRed}{\textbf{0.76}} & 0.72 & 0.55 \\
Spi-Net \cite{gesnouin2020predicting} & Skelton & \color{ForestGreen}{\textbf{0.59}} & 0.61 & 0.71 & 0.50 \\
TrouSPI-net \cite{gesnouin2021trouspi} & Skelton & \color{ForestGreen}{\textbf{0.59}} & \color{OrangeRed}{\textbf{0.76}} & 0.56 & 0.32 \\ \hline
\multirow{5}{*}{Our Method} & \multicolumn{1}{l|}{B} & 0.50 & 0.63 & 0.67 & 0.51 \\
 & \multicolumn{1}{l|}{B+L} & 0.51 & 0.69 & 0.69 & 0.57 \\
 & \multicolumn{1}{l|}{B+L+G} & 0.53 & 0.75 & 0.82 & 0.71 \\
 & \multicolumn{1}{l|}{B+L+G+P} & 0.52 & 0.75 & 0.83 & 0.72 \\
 & \multicolumn{1}{l|}{B+L+G+P+M} & \color{OrangeRed}{\textbf{0.60}} & \color{ForestGreen}\textbf{0.75} & \color{ForestGreen}{\textbf{0.85}} & \color{ForestGreen}\textbf{0.73} \\ \hline
\end{tabular}
\label{exp}
\end{table}

We examine different combinations of input features as an %abolition 
ablation study in Table \ref{exp}. These features are depicted in Figure \ref{results} for two different traffic scenarios. Our baseline method relies on a single feature and uses only bounding box coordinates. It reaches the lowest rate of AUC and F1-score among the other methods. A significant improvement by ~6\% and ~3\% in AUC and F1 is achieved by adding the motion features to the baseline method.
Adding semantic information to parse the global context of the scene increased the AUC and F1 score by 12.5\% with respect to the baseline method. Our initial experiments on extracting global contextual information from the scene indicated that considering more traffic-related objects such as other road users and vehicles results in an increase in intention classification accuracy. For example, using only road segmentation instead of the road and  road users segmentation could decrease the AUC by 1.5\% and F1 by 1.8\%.

The experiments reveal the combination of the pedestrian's bounding boxes coordinates, body joints, spatial features, semantic information, and motion features, leads to the best results on the overall AUC and F1 score combined for both JAAD$_{beh}$ and JAAD$_{all}$. Also, our proposed method could achieve the highest AUC for JAAD$_{beh}$.
\vspace{-1mm}
In addition, the results obtained from our preliminary experiments show that the outcome of the model highly depends on the starting point and moment of the analysis. This is possibly due to the complexity of the environment and also the occasions of partial and fully occluded pedestrians in the scene.  This is in line with the claims by \cite{Lorenzo2021CAPformerPC} and \cite{gesnouin2022assessing}.
We also observed that using an accurate pose estimator and scene objects/motion descriptor (e.g., semantic segmentation, optical flow, etc.) leads to outperforming results on the currently available datasets. 
\vspace{-1mm}
Figure \ref{results} illustrates the extracted features among two different video samples of the JAAD dataset and the final prediction of our model in intention classification.
The model reports 96\% and 87\% confidence rates of classification for the samples shown in Figure \ref{out1} and \ref{out2}, respectively.

\section{Conclusion}\label{conclusion}

We introduced the fusion of various pedestrian and scene features for a better prediction of the pedestrian's crossing intention. The study showed that assessing pedestrian behaviour based on single individual features such as the pedestrian's location, body pose, or global context of the environment will lead to marginally lower performance than the proposed model due to the absence of other complementary influencing factors. It is therefore concluded that relying on multiple features including appearance, pedestrian pose, and vehicle motion, as well as the surrounding information produces better results.
We also hypothesised that utilisation of the local and global spatio-temporal features helps to better understand and predict pedestrian crossing intention. 
Our experiments showed that although the proposed method is not the best from all aspects, it is the superior model overall, achieving the best results in combined AUC and F1-score among 15 other models on JAAD dataset metrics.

The majority of current studies including this research, train their  intention prediction model based on the AV's point of view videos. This is while pedestrians may change their decisions also under the influence of AV's speed, distance, lane, and manoeuvre behaviour. In other words, the pedestrians' point of view has been widely neglected. Therefore, the integration of ego-vehicle kinematics into the prediction models can be a sensible approach for future studies. 

\section*{Acknowledgement} 
The authors would like to thank all partners within the Hi-Drive project for their cooperation and valuable contribution. This research has received funding from the European Union's Horizon 2020 research and innovation programme, under grant agreement No. 101006664. The article reflects only the authors' view and neither European Commission nor CINEA is responsible for any use that may be made of the information this document contains.

 \bibliographystyle{splncs04}
\bibliography{ref.bib}

\end{document}